\documentclass[Afour,sageh,times]{sagej}

\usepackage{moreverb,url}
\usepackage{amsmath,mathtools}
\usepackage{amssymb}
\usepackage{multirow}
\usepackage{colortbl,booktabs}
\usepackage{array,ragged2e}
\usepackage{textcomp} 
\usepackage{graphicx,subcaption,ragged2e}
\usepackage{graphicx}
\usepackage{caption}
\usepackage{url}
\usepackage{graphics}
\usepackage[T1]{fontenc} 
\usepackage{floatrow}
\newfloatcommand{capbtabbox}{table}[][\FBwidth]
\usepackage{blindtext}

\usepackage{array}
\newcolumntype{W}{>{\centering\arraybackslash}m{5mm}}

\usepackage[colorlinks,bookmarksopen,bookmarksnumbered,citecolor=red,urlcolor=red]{hyperref}
\definecolor{lightgray}{rgb}{0.767,0.767,0.767}

\usepackage{xcolor}
\definecolor{irblue}{HTML}{0071bc}
\definecolor{irorange}{HTML}{d85218}
\definecolor{iryellow}{HTML}{ecb01f}
\definecolor{irpurple}{HTML}{7d2e8d}
\definecolor{irgreen}{HTML}{76ab2f}
\definecolor{irbrown}{HTML}{a1132e}
\definecolor{irred}{HTML}{999999}

\newcommand\BibTeX{{\rmfamily B\kern-.05em \textsc{i\kern-.025em b}\kern-.08em
T\kern-.1667em\lower.7ex\hbox{E}\kern-.125emX}}

\begin{document}
\setcounter{secnumdepth}{3}

\runninghead{MassMIND: Massachusetts Maritime INfrared Dataset}

\title{MassMIND: \textbf{Mass}achusetts \textbf{M}aritime \textbf{IN}frared \textbf{D}ataset}

\author{Shailesh Nirgudkar\affilnum{1}, Michael DeFilippo\affilnum{2}, Michael Sacarny\affilnum{2}, Michael Benjamin\affilnum{3} and Paul Robinette\affilnum{1}}

\affiliation{\affilnum{1}University of Massachusetts, Lowell, MA, USA\\
\affilnum{2}MIT Sea Grant, Autonomous Underwater Vehicles Lab, MA, USA\\
\affilnum{3}Massachusetts Institute of Technology Laboratory for Autonomous Marine Sensing Systems, Cambridge, MA, USA}

\corrauth{Shailesh Nirgudkar,
University of Lowell,
Lowell, MA, USA 01854}

\email{shailesh\_nirgudkar@student.uml.edu}

\begin{abstract}
Recent advances in deep learning technology have triggered radical progress in the autonomy of ground vehicles. Marine coastal Autonomous Surface Vehicles (ASVs) that are regularly used for surveillance, monitoring and other routine tasks can benefit from this autonomy. Long haul deep sea transportation activities are additional opportunities. These two use cases present very different terrains - the first being coastal waters- with many obstacles, structures and human presence while the latter is mostly devoid of such obstacles. Variations in environmental conditions are common to both terrains. Robust labeled datasets mapping such terrains are crucial in improving the situational awareness that can drive autonomy. However, there are only limited such maritime datasets available and these primarily consist of optical images. Although, Long Wave Infrared (LWIR) is a strong complement to the optical spectrum that helps in extreme light conditions, a labeled public dataset with LWIR images does not currently exist. In this paper, we fill this gap by presenting a labeled dataset of over 2,900 LWIR segmented images captured in coastal maritime environment under diverse conditions. The images are labeled using instance segmentation and classified in seven categories - sky, water, obstacle, living obstacle, bridge, self and background. We also evaluate this dataset across three deep learning architectures (UNet, PSPNet, DeepLabv3) and provide detailed analysis of its efficacy. While the dataset focuses on the coastal terrain it can equally help deep sea use cases. Such terrain would have less traffic, and the classifier trained on cluttered environment would be able to handle sparse scenes effectively. We share this dataset with the research community with the hope that it spurs new scene understanding capabilities in the maritime environment. 
\end{abstract}


\maketitle
\section {INTRODUCTION}
In recent years, advances in deep learning algorithms have led to an exponential growth in the research on autonomy of land vehicles. Some of the key catalysts for this growth have been publicly available labeled datasets, open-source software, publication of novel deep learning architectures and increase in hardware compute capabilities. The maritime environment, with an abundance of periodic tasks such as monitoring, surveillance, and long haul transportation presents a strong potential for autonomous navigation. Availability of good datasets is a key dependency for gaining autonomy. Various types of sensors such as electro-optical (EO) cameras, LWIR cameras, radar and lidar help in collecting huge volume of data about surroundings. The challenge lies in interpreting this data and creating labeled datasets that can help train deep learning architectures. 

EO cameras are predominantly used to capture images because of their versatility and the abundant Convolutional Neural Network (CNN) architectures \cite{cnn} that learn from labeled images. Two methods are commonly used for annotating an image. First, detecting objects of interest by drawing bounding boxes around them. Second, semantically segmenting an image in which each pixel is assigned a class label. The first method is faster as it focuses on specific targets, however, the second is more refined as it segments the entire scene. 

\begin{figure}
\setlength{\tabcolsep}{1pt}
\centering
\begin{tabular}{cc}
\subfloat[Optical in sun glare \hfill]{\includegraphics[width=1.55in]{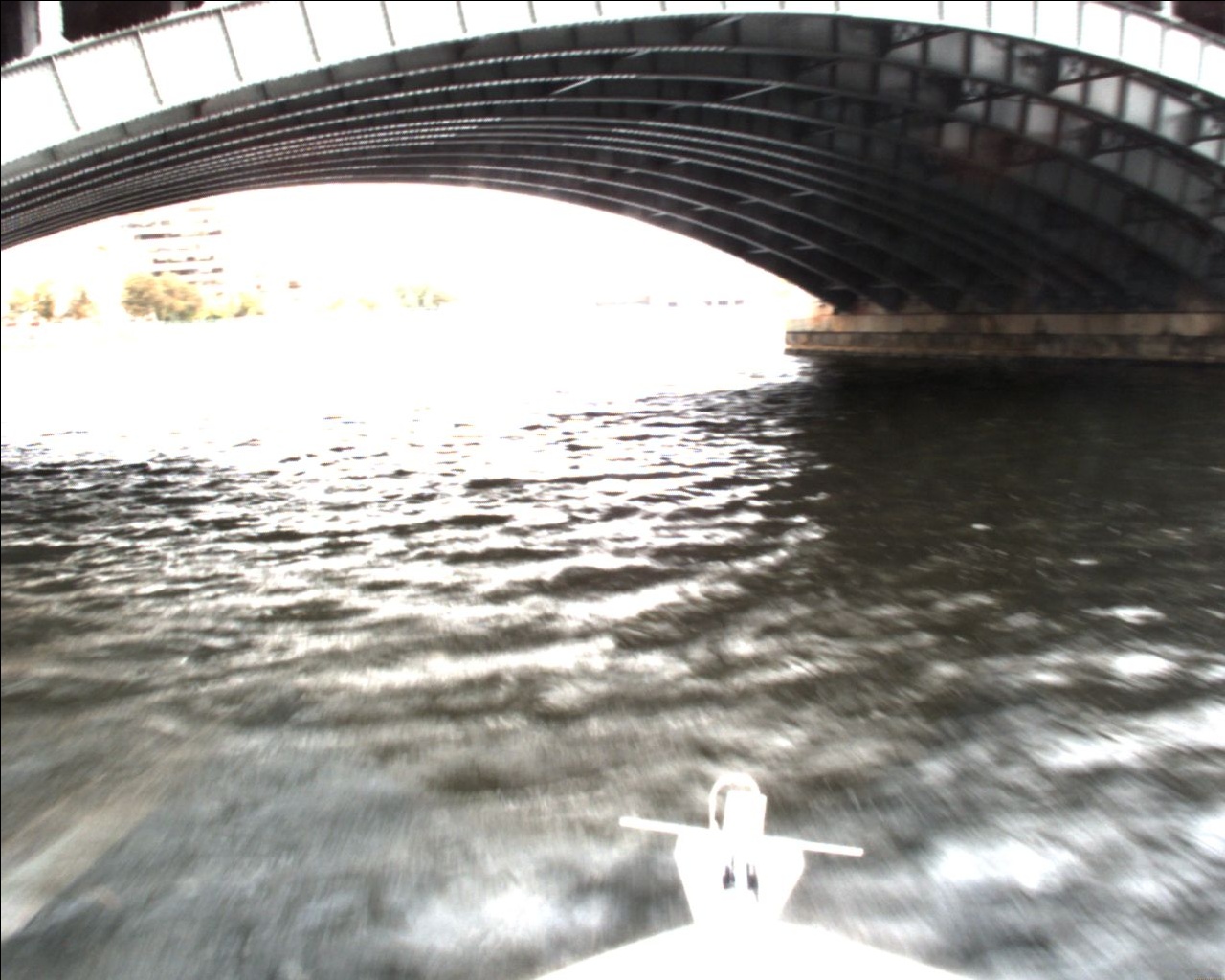}} &
\subfloat[LWIR in sun glare]{\includegraphics[width=1.55in]{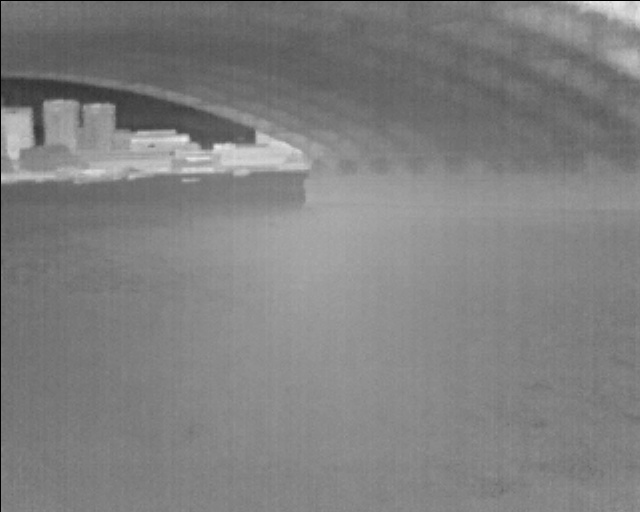}}\\
\subfloat[Optical in dark]{\includegraphics[width=1.55in]{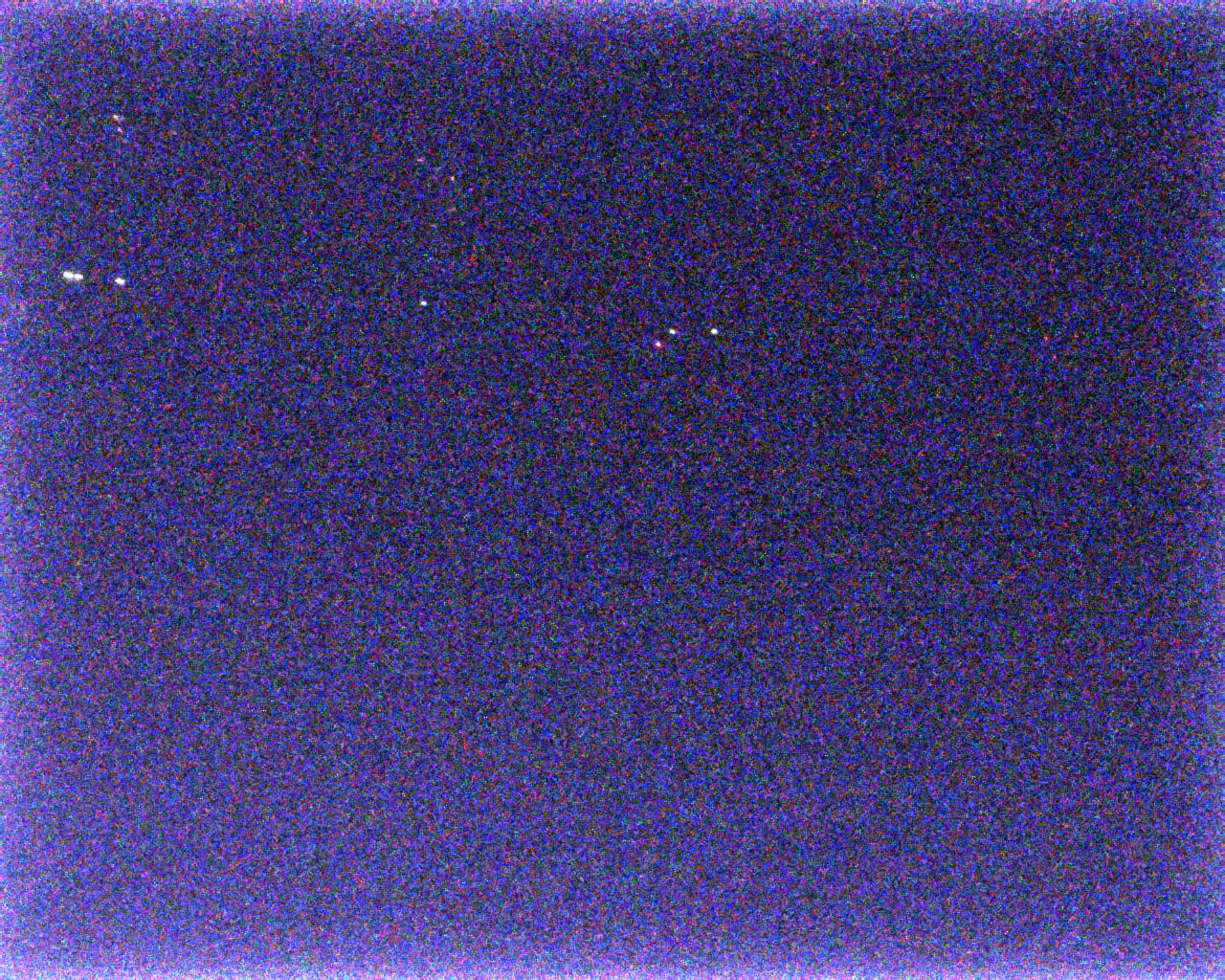}} &
\subfloat[LWIR in dark]{\includegraphics[width=1.55in]{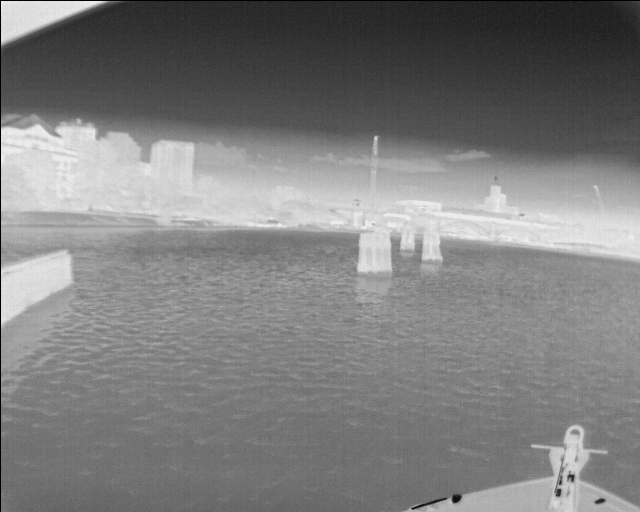}} \\
\end{tabular}
\caption{LWIR vs Optical cameras in extreme light conditions: In (a), sun's glare is obstructing the view for optical while the visibility is clear in LWIR (b). In (c) visibility from optical camera is almost zero in the dark, whereas LWIR continues to give a clear 
picture (d).}
\label{optical_ir_comparison}
\end{figure}

The maritime environment is predominantly exposed to sky and water and the lighting conditions are therefore quite different as compared to a ground terrain. Glitter, reflection, water dynamism and fog are common. These conditions deteriorate the quality of optical images. Horizon detection is also a common problem faced while using optical images. On the other hand, LWIR images offer distinct advantages in such extreme light conditions as shown experimentally by \cite{sensor_evaluation_robinette}, \cite{shailesh_icar2021} and as shown in Fig. \ref{optical_ir_comparison}. Marine robotics researchers have used LWIR sensors in their work (\cite{scholler}, \cite{rodin}, \cite{qi}, \cite{wang}). Most notably \cite{vais} has built a labeled dataset of paired visible and LWIR images of different types of ships in the maritime environment. However, there are certain limitations of this dataset which are discussed in the next section. 

\par In this paper, we present a dataset of over 2,900 LWIR maritime images that capture diverse scenes such as cluttered marine environment, construction, living entities and  near shore views across various seasons and times of the day in the Massachusetts Bay area including Charles River and Boston Harbor. The dataset images are segmented using instance and semantic segmentation into 7 classes. We also evaluate the performance of this dataset across 3 popular deep learning architectures (UNet, PSPNet, DeepLabv3) and present our findings with respect to obstacle detection and scene understanding. The dataset is publicly available and can be downloaded from the URL \cite{our_dataset}. We aim to stimulate research interest in the field of perception in maritime autonomy through these dataset.   
\par The remainder of the paper has been organized as follows. Section 2 outlines the current state of art in the maritime domain. Section 3 describes the hardware assembly that is used for data acquisition. Specifics about the dataset and segmentation methods have been elaborated in Section 4 and Section 5 presents  evaluation results against the 3 architectures. We draw conclusions in Section 6 and highlight the future plan of work.

\section {RELATED WORK}
Significant research has gone in the field of marine robotics over the past few years and researchers have developed multiple datasets that can aid autonomy. These datasets vary in size and complexity, however, their main purpose revolves around environment monitoring and surveillance. \cite{intcatch} used autonomous aquatic drones and captured multivariate time series data for the purpose of water quality management. The dataset contains parameters such as water conductivity, temperature, oxygen level and is useful for time series analysis. \cite{mardct} used ASV in Venice, Italy and created a dataset of optical images for the purpose of boat identification and classification. The ground truth in this dataset is provided in the form of bounding boxes. \cite{seagull} captured images from an unmanned aerial vehicle (UAV) 150-300 meter above ocean surface for research on sea monitoring and surveillance. Though multi-spectrum cameras including IR cameras are used, the dataset is not annotated and hence it cannot be used readily for model training. Another problem is that the images were aerial pictures and therefore do not have the same perception properties as that of the view from a vehicle on the water surface. \cite{vais} has created a dataset of paired visible and LWIR images of various ships in the maritime environment. The annotations are provided in the form of bounding boxes around ships. However, selection is focused on those images where ships occupy more than 200 pixels of the image area. The dataset is therefore not adaptable for general use where obstacles can be in any shape or form such as buoys, small platforms, construction in water. More importantly, the images are captured from a static rig mounted on the shore and therefore lack the dynamism found in real life scenarios. 
\par \cite{smd} has created a dataset using optical and near infrared (NIR) cameras to capture maritime scenes from a stationary on-shore location. This dataset also has some video sequences recorded from a ship but they are from a higher vantage point and hence do not clearly depict the same view as that from an ASV. Additionally, not all images in the dataset are annotated. Only a few are annotated using bounding boxes. \cite{ipatch} has used multi-spectrum cameras to capture video segments of enacted piracy scenes. The emphasis is on threat detection and so the video sequence contains medium to large boats appearing in the frames. Since the visuals are enforced, the dataset lacks natural marine scenes that are critical for training a model. \cite{intcatch2} used pixel wise segmentation to differentiate between water and non-water class. The dataset contains around 515 optical images with the 2 classes. In recent years, two excellent datasets MaSTR1325 and MODS have been created by \cite{mastr} and \cite{mods}. The optical images in these datasets are semantically segmented in 3 classes and are suitable for unmanned surface vehicles (USV) navigation. However, as pointed out by \cite{mods}, bright sunlight, foam, glitter and reflection in the water deteriorate the inference quality. Small object detection poses a challenge even for the model developed by \cite{wasr} that is specifically designed for the maritime environment. 
\par
\begin{figure}[t]
\begin{center}
\includegraphics[width=0.8\textwidth]{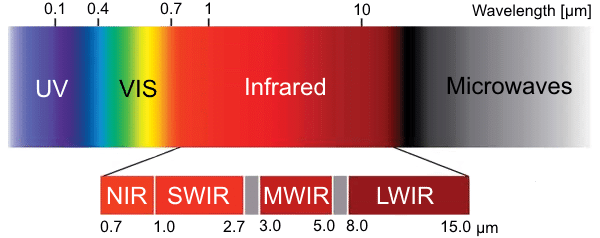}
\caption{IR Spectrum}
\label{IRSpectrum}
\end{center}
\end{figure}

 Most of these datasets are captured using EO cameras. The ones that used NIR, do not provide the benefit of LWIR as they are still dependent on visible light. As indicated in Fig. \ref{IRSpectrum} \cite{edmund_optics} NIR light occupies a wavelength between 0.75 to 1.4 microns and although it is a part of the Infrared spectrum, it is not caused by thermal radiation. Instead it is almost similar to visible light except that it cannot be observed by human eye. Heat or thermal radiation is the primary source of radiation for LWIR  \cite{nasa_thermal}, \cite{nir_vs_lwir_compare} and therefore LWIR is truly independent of visible light.
\par In a nutshell, the existing NIR/LWIR datasets lack complete scene parsing capabilities, or have been created in a controlled environment or are not publicly available. MassMIND bridges this gap and is the largest labeled dataset of its kind containing real-life LWIR images. The images have been recorded in a busy marine area as well as in not so busy ocean waters during various seasons and times of the day over a period of 2 years. The dataset has been annotated into 7 classes using instance segmentation that can be used for scene parsing during autonomous navigation.   

\section {SYSTEM SETUP}
\begin{figure}[t]
\begin{center}
\includegraphics[width=1.0\textwidth]{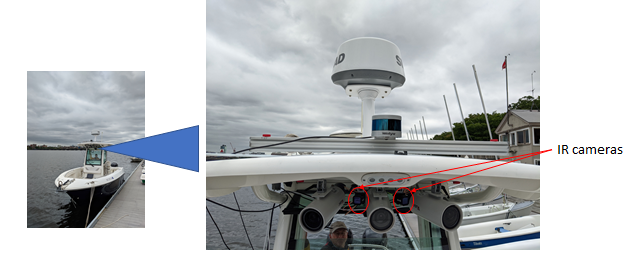}
\caption{R/V Philos showing placement of LWIR cameras}
\label{philos_system}
\end{center}
\end{figure}

R/V Philos, an ASV owned by the Brunswick Corporation and operated by MIT Autonomous Underwater Vehicle (AUV) \cite{auv} lab was instrumental in collecting images for the MassMIND dataset. The ASV is equipped with 3 EO cameras, 2 Forward looking Infrared (FLIR)-LWIR cameras, a radar, a lidar, GPS and Inertial Measurement Unit (IMU). The FLIR cameras used in 2019 were entry level providing resolution of 320$\times$240 (width$\times$height) and horizontal field of view of 34\textdegree. These cameras were upgraded in 2020 which provided resolution of 640$\times$512 (width$\times$height) and horizontal field of view of 75\textdegree. Images from these sensors have slight offset with one another due to the placement of EO and FLIR cameras as shown in Fig. \ref{philos_system}. Table \ref{lwir_spec} shows technical specifications of the FLIR ADK cameras used in the setup in 2019 and later. Additional details of the ASV can be found in \cite{robowhaler}. The unannotated raw dataset from the sensors was also released as part of that work \cite{unannotated_dataset}.

\begin{table}[th]
\caption{FLIR ADK Data sheet}
\label{lwir_spec}
\centering
\small
\resizebox{0.95\columnwidth}{!}{%
\begin{tabular}{lll}
\hline
\rowcolor{lightgray}\textbf{Thermal Imager} & \textbf{FLIR ADK, >2019} & \textbf{FLIR ADK, 2019}\\
\hline
Data Format & 16 bit TIFF or compressed & 16 bit TIFF \\
 & 8 bit PNG &  \\
\hline 
Frame Rate & 30 Hz/60 Hz selectable, & Same \\
& 9 Hz optional  \\
\hline
Spectral Band & 8-14 \textmu m (LWIR) & Same \\ 
\hline
Array format & 640 $\times$ 512 & 320 $\times$ 256 \\
\hline
Field of View & 75\textdegree & 34\textdegree \\
\hline
Pixel pitch & 12 \textmu m & Same \\
\hline
Thermal Sensitivity & < 50 mK & Same \\
\hline
\rowcolor{lightgray}\textbf{Power} &  &\\
\hline
Power Consumption & 1W (without heater), & 580 mW \\
 & 4W average, \\
 & 12W maximum (with heater) \\
\hline
\rowcolor{lightgray}\textbf{Environmental}& &\\
\hline
Operating Temperature & -40 \textdegree C to 85 \textdegree C & Same\\
\hline
Environmental Protection & IP67 & Same\\
\hline
Shock & 1500 g @ 0.4 msec & Same\\
\hline
\rowcolor{lightgray}\textbf{Physical} & & \\
\hline
Dimensions (W $\times$ H $\times$ D) & 35 $\times$ 40 $\times$ 47 mm &  38 $\times$ 38 $\times$ 42.5 mm\\
\hline
Weight & 100 gm & 116.12 gm \\
\hline
\end{tabular}%
}
\end{table}

\section {DATASET DESCRIPTION}
The strength of LWIR is its independence from visible light. Our main goal was to utilize this property of LWIR and create a comprehensive marine dataset of real-life images so that it complements existing optical datasets. In this section, we first describe how the data was acquired and the process of image selection. Then we describe the segmentation scheme that was followed along with the labeling process.  

\begin{figure}[htbp]
\begin{center}
\includegraphics[width=1.0\textwidth]{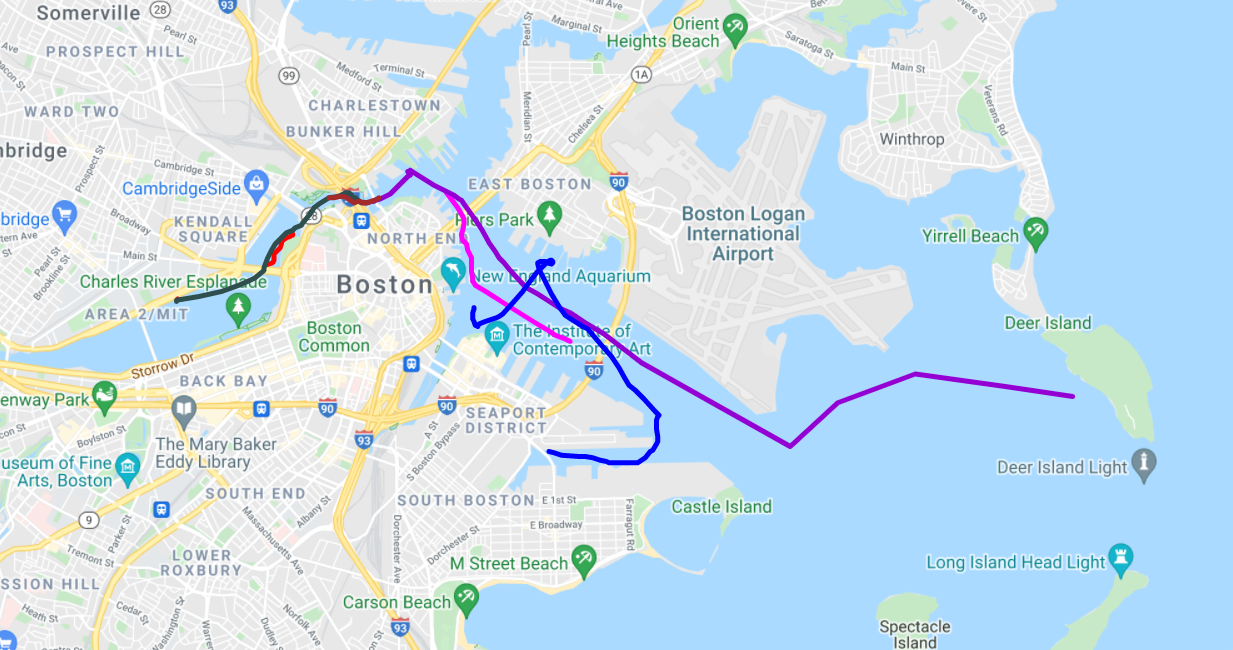}
\caption{Area covered by R/V Philos near Boston, Massachusetts }
\label{map}
\end{center}
\end{figure}

\subsection {Data Acquisition and Selection}
R/V Philos recorded scenes around the Massachusetts Bay area over a period of 2 years from 2019 till November 2021. This gave us an opportunity to create seasonal and temporal diversity in the dataset. Nearshore scenarios were also captured as they present more challenging scenes with construction and increased human and vehicle presence. This was in addition to the complexity present due to weather and light patterns observed in the farther seas. Images were taken in a natural, uncontrolled environment which makes the dataset applicable for practical use. Fig. \ref{map} shows some of the typical routes the ASV has taken in these trips.
\par R/V Philos stores data from various sensors in the form of rosbag \cite{rosbag} files. LWIR and optical images were extracted from these rosbag files and carefully analyzed. One challenge faced while selecting LWIR images was the lack of texture and color information that sometimes undermined the scene complexity. In such cases, we used the corresponding optical equivalents to guide through the selection. The epoch time from the rosbag file was also used to specifically target scenes after sunset. This ensured that our dataset specifically included scenes that would pose challenges to the optical sensors. The resolution of 2019 footage was lesser than later recordings and helped create a variation that would be natural in a practical world. The dataset has approximately equal distribution of these images as indicated in Table \ref{image_statistic}. 
\par
\begin{figure}[t]
\begin{center}
\includegraphics[width=8.6cm,height=5cm]{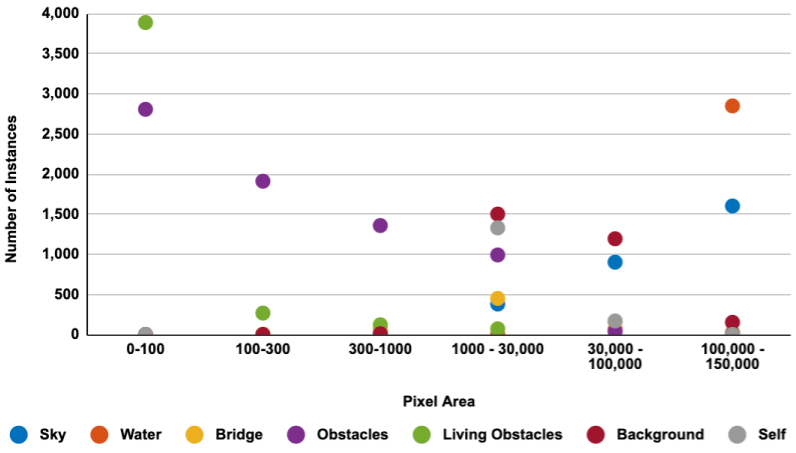}
\caption{Distribution of instances across pixel area: As seen, sky, water and background are prominently on the right end of the chart with larger area. Most of the concentration of living obstacles and obstacles is on the left indicating smaller pixel areas. There is a gradual progression toward the center which indicates cases where these obstacles are in close proximity and appear larger. Distance does play a role in bridge size as well but not to the same extent as that of obstacles. Training was therefore done on actual size images so that we can retain information for obstacles and living obstacles.}
\label{instance_concentration}
\end{center}
\end{figure}

 Intrinsically maritime scenes have a significant imbalance in their class distribution as compared to the ground scenes. Sky and water are present in almost any image and occupy major portions. While nearshore scenes help in capturing the real life traffic, and living presence, they also amplify the background adding to the disproportion. As part of the selection process, we tried to reduce this imbalance so that sky and background do not undermine the obstacles and living obstacles in the water that are critical for training. We also focused on bridges as a part of them was relevant for path navigation while the remainder was not and therefore had to be considered as background. Another important difference between maritime and ground scenes is the line of sight. In maritime environment, cameras are exposed to large vistas and distant objects although part of the scene, can only be identified vaguely because they appear tiny. Annotation of such objects may vary as they become more discernible. Such reinterpretation requires consideration of distance as an additional parameter in the classifiers. Table \ref{size_variation} and Fig. \ref{instance_concentration} show the size and instance distribution of each class in the dataset. The obstacles (purple) and living obstacles (green) mostly occupy less area but there are few instances in which they occupy considerably larger area. The images in such cases contain close up views of boat and/or humans. Fig. \ref{instance_concentration} shows size variation across far obstacles, sky and water in the order of magnitude.

\begin{table}[t]
\caption{Image distribution: 2019 sensors had low resolution as compared to 2020}
\label{image_statistic}
\begin{center}
\footnotesize
\begin{tabular}{cc}
\hline
\rowcolor{lightgray}\textbf{Year of recording} & \textbf{Number of images}  \\
\hline
2019 & 1423 (48.8\%)  \\
\hline 
> 2019 & 1493 (51.2\%) \\
\hline
\rowcolor{lightgray}\textbf{Total} & \textbf{2916} \\
\hline
\end{tabular}
\end{center}
\end{table}

\begin{table}[t]
\caption{Variation of pixel area across classes: As expected, sky, water and background are generally larger in most of the images. Obstacles and living obstacles on the other hand vary a lot based on the distance. Nearer ones contribute to much larger areas than the distant ones. Pixel area of 2 represents very small buoys or birds.}
\label{size_variation}
\begin{center}
\footnotesize
\begin{tabular}{ccc}
\hline
\rowcolor{lightgray}\textbf{Class} & \textbf{Minimum pixel area} & \textbf{Maximum pixel area }  \\
\hline
Sky & 303 & 235,763  \\
\hline 
Water & 3,013 & 251,531 \\
\hline
Bridge & 29 & 223,990 \\
\hline
Obstacles & 2 & 17,341 \\
\hline
Living Obstacles & 2 & 12,512 \\
\hline
Background & 48 & 249,720 \\
\hline
Self & 32 & 11,915 \\
\hline
\end{tabular}
\end{center}
\end{table}

\subsection {Image Segmentation}
We analyzed the Cityscape \cite{cityscapes} dataset to understand the guidelines of their classification. In order to make our dataset meaningful for path navigation, we finalized 7 classes - `sky', `water', `obstacle', `living obstacle', `bridge', `background' and `self' for annotation.
Categorizing maritime scene into such refined classes ensures that true obstacles are identified and not get combined with other non-essential entities. Segmentation guidelines were prepared to assist the annotators. The `obstacle' class consists of inanimate objects surrounded by water such as buoys, small platforms, sailboats, ships, kayaks. Animate obstacles like humans and birds in the water are classified as `living obstacles'. Identification of living entity may be useful for navigation such as deriving COLREGs compliant navigation path \cite{colreg} and also for other use cases such as surveillance and environmental or ecological monitoring. Because of the urban setting of the dataset, bridges were frequent and posed a unique challenge. The legs of the bridge that are submerged in water and the belly of the bridge were relevant as the ASV would need to maneuver around the same. However, the top of the bridge or construction on top was not. We decided to annotate the navigation relevant portions of the bridge as `bridge' class and mark the remaining portions as belonging to the `background' class. The `background' class covered any static or dynamic entities that were on land such as trees, buildings, construction cranes and other structures. We observed self returns on the images due to the position of the cameras and these have been labeled as `self' class. This area neither represents `obstacle' nor `background' and hence justified its own label. 
The path planning module can always be programmed to exclude the `self' class, but it helps to identify the regions in the image correctly. Table \ref{class_distribution_table} shows number of instances in each class that were annotated in these images and also the distribution of pixel area across these classes.  
\par As we were working through the annotations, many images gave us new insights that had to be incorporated. We had anticipated bridge complexity as part of our planning, however we encountered a similar issue with respect to obstacles. Initially we defined an obstacle as an entity surrounded by water on all sides. However, we came across construction objects that stretched from land into the water such as gates that allow the boats to pass through, piers and docks to park the boats, or platforms connected to the land. These obstacles were pertinent for path navigation and hence could not be treated as background. Similarly birds flying just above the water were categorized as sky, and birds in the water were categorized as `living obstacle'. Since the images are 2D and LWIR does not give a good feel of the color and texture, there were cases where the obstacle in water was collapsed in the rear background and had to be reviewed thoroughly. Another challenge faced was due to the wide and deep line of sight that we get in the marine environment. There were many obstacles near the horizon that were very tiny. We had to enlarge the image and annotate such images per pixel. A person on a boat is annotated as a `living obstacle' while the boat is annotated as an `obstacle'. In cases where the boat was far out and a human could not be easily identified on the boat, it was not called out and considered as part of the `obstacle' class. 

\begin{table}[t]
\begin{center}
\caption{Spread of each class in the dataset: As seen all classes have a proportionate presence in the dataset. Due to instance segmentation, multiple occurrence of the same class in an image is treated as a new instance. The dataset has a significant number of obstacles and living obstacles - entities impacting the navigation path. They are smaller and their area is far less than that of the dominant classes like sky and water. Self class does appear in considerable images and was retained as the model can learn from it. Background, sky and self are always counted as one even if they appear multiple times in an image.}

\label{class_distribution_table}
\footnotesize
\begin{tabular}{m{16mm}  m{12mm}  m{15mm} m{15mm} m{14mm}} \hline
\rowcolor{lightgray}\textbf{Class} & \textbf{Total number of instances} & \textbf{\% Distribution of number of instances} & \textbf{\% Distribution of pixel area} \\  \hline
Sky & 2902 & 12.97 & 30.58 \\ \hline
Water & 2916 & 13.04 & 52.21 \\ \hline
Bridge & 715 & 3.19 & 1.67 \\ \hline
Obstacles & 7120 & 31.83 & 0.94 \\ \hline
Living Obstacles & 4350  & 19.45  & 0.05\\ \hline
Background & 2860 & 12.78 & 11.28 \\ \hline
Self & 1501  & 6.71 & 3.25 \\ \hline
\end{tabular}
\end{center}
\end{table}

\par
\begin{figure}
\setlength{\tabcolsep}{1pt}
\centering
\begin{tabular}{cc}
\subfloat{\includegraphics[width=1.55in]{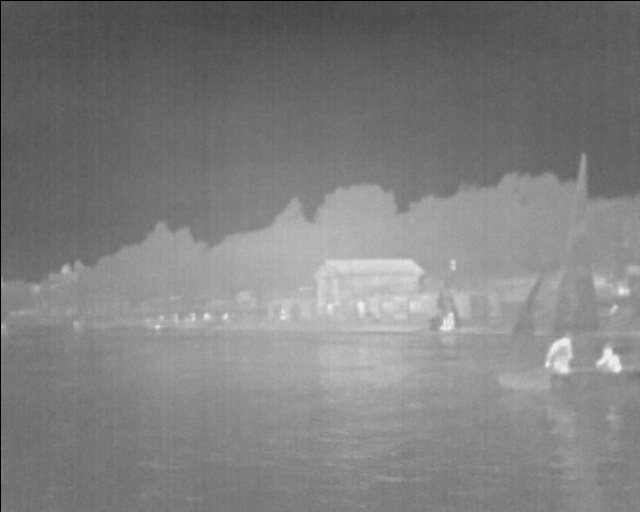}} &
\subfloat{\includegraphics[width=1.55in]{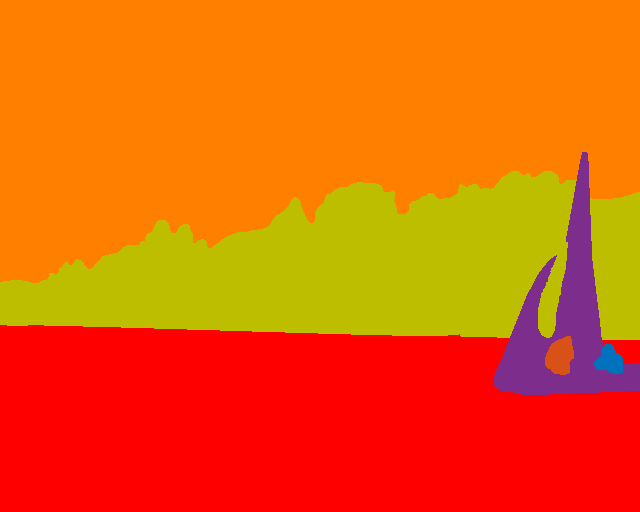}} \\
\subfloat{\includegraphics[width=1.55in]{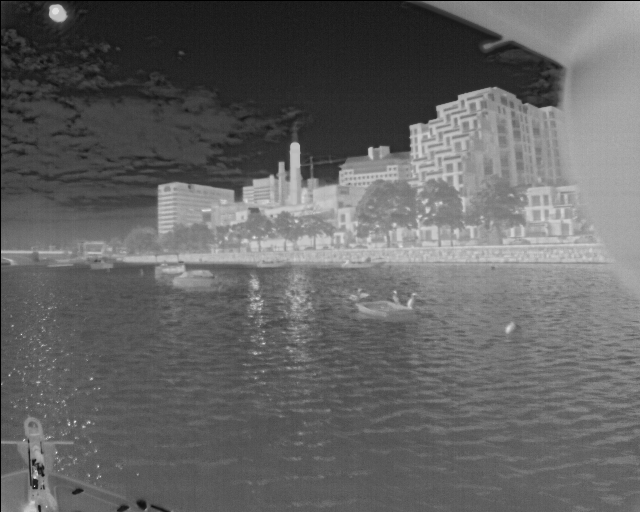}} &
\subfloat{\includegraphics[width=1.55in]{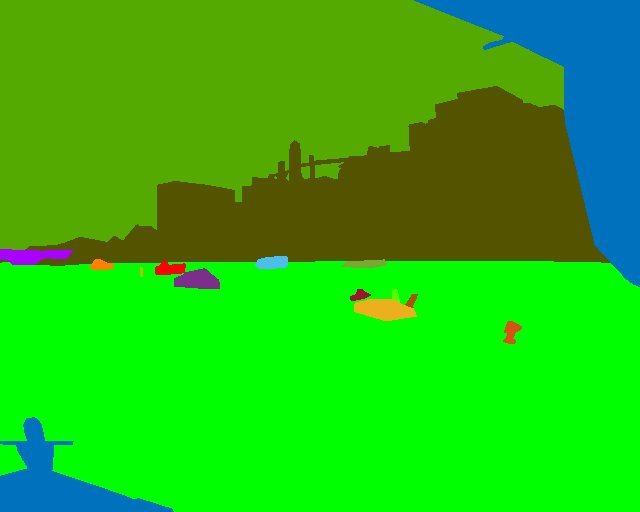}} \\
\end{tabular}
\caption{LWIR images and their segmentation: Multiple occurrences of birds, humans or boats have their own instance id and appear in different color. Background, sky and water each have only one instance in an image.}
\label{segmentation_work}
\end{figure}

 In addition to labeling, we also had to make a choice between semantic and instance segmentation. In semantic segmentation, each occurrence of a class is assigned same class label while in instance segmentation, each instance of a class is assigned different instance id. Unlike on land, maritime environment has a wide and deep line of sight. The near and far objects may overlap in the 2-D view of an image. They will be collapsed if semantic segmentation was used. By instance segmenting the image, the information of overlapping multiple obstacles can be retained which in turn can be used to evaluate or develop occlusion aware classifier vital for path planning. Lastly, the benefit of
instance segmentation is that it can always be converted programmatically to semantic based on the need.
Fig. \ref{segmentation_work} illustrates a few samples of raw LWIR images and their annotated masks. 

\subsection {Labeling Process}
\begin{figure}[t]
\begin{center}
\includegraphics[width=0.75\textwidth]{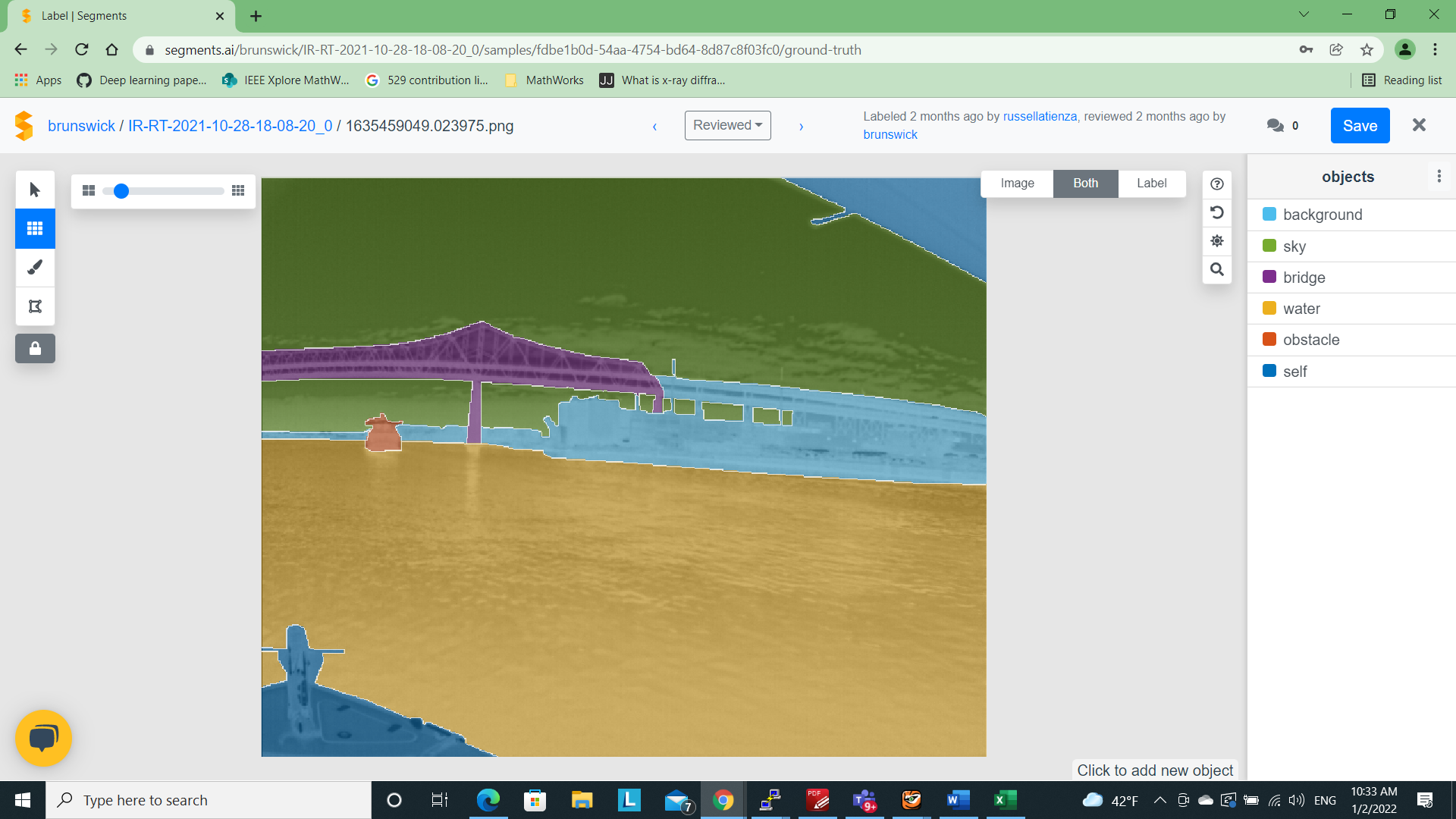}
\caption{Image segmentation platform: Segments.ai tool that was used for annotation. It provides super pixels that improve the productivity and reduces manual errors. Pipeline was automated using the API interface.}
\label{segments_screen}
\end{center}
\end{figure}

An independent organization was tasked with the labeling of these images. To achieve consistency in the result, we created and shared a detailed class label definition document so that labelers would have a common understanding. We also labeled a batch of images ourselves to illustrate the desired outcome. There are many tools available for image segmentation but we chose \cite{segments} platform because of its ease of use. The labeling was quick because of it's superpixels technology. The API interface made working with the labeled dataset efficient. Fig. \ref{segments_screen} shows the tool in action.
While the labeling tool was easy to use, there were some challenges faced especially due to the nature of LWIR images. Lack of color and textures posed difficulties in image identification as well as annotation. Color and texture assists the human eye in making a judgment. Lack of thereof, led to multiple reviews in some complex situations such as cluttered set of obstacles, cluster of parking poles, flock of birds or distant ships. For each LWIR video, the corresponding optical version was therefore provided to the labeling team for reference. The image had to be classified in 7 classes. This made the labeling process quite elaborate and time consuming. Identifying parts of the bridge that should be annotated as background was subjective and as a result, there may be some deviations in some of the annotations. In summary, the semantic annotation exercise for LWIR images was more time consuming and needed a lot more attention than we had anticipated.

2 rounds of reviews were done to improve the accuracy of segmentation. Including the reviews, each image annotation took approximately 25 minutes. We also developed a script to check for any unclassified pixels and corrected such masks. Automated checks for validating the segmentation and gathering metrics were also developed.

\section {DATASET EVALUATION}
Though we have created instance-segmentation dataset, for the purpose of evaluation we used its semantic segmentation masks. To evaluate the efficacy of the dataset, we used 3 image segmentation architectures - UNet developed earlier by \cite{unet} for medical image segmentation, PSPNet by \cite{pspnet} for its improvement over fully convolutional network based segmentation and DeepLabv3 by \cite{deeplab} which is known for its overall superior performance. 
The dataset was split into 3 categories - 70\% for training, 20\% for validation and 10\% for testing. The test images were used only during inference. Pretrained weights such as \cite{imagenet} are generally available for training optical images. They greatly reduce the time to retrain the model, however these do not work well for LWIR image training as shown by  \cite{shailesh_icar2021}. We therefore trained these architectures from scratch for our LWIR dataset. Original image resolution of 640$\times$512(width$\times$height) pixels was used across all the architectures - UNet  (\cite{unet}), PSPNet (\cite{pspnet}), DeepLabv3 (\cite{deeplab}) for both 2019 as well as the 2020 images. 

\subsection {Data Augmentation}
\begin{figure}
\setlength{\tabcolsep}{1pt}
\centering
\begin{tabular}{ccccc}
\subfloat{\includegraphics[width=0.19\textwidth]{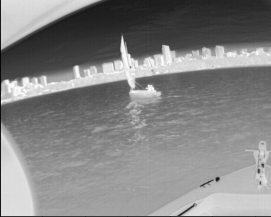}} &
\subfloat{\includegraphics[width=0.19\textwidth]{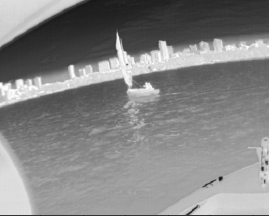}} & 
\subfloat{\includegraphics[width=0.19\textwidth]{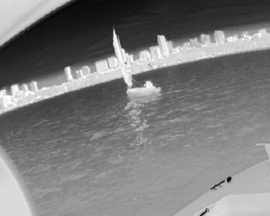}} & 
\subfloat{\includegraphics[width=0.19\textwidth]{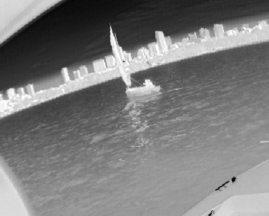}} & 
\subfloat{\includegraphics[width=0.19\textwidth]{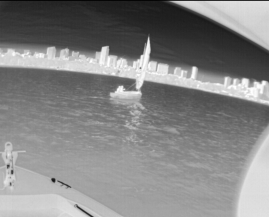}}  \\
\subfloat{\includegraphics[width=0.19\textwidth]{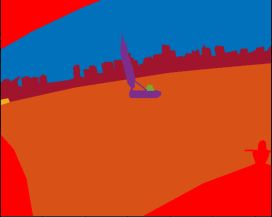}} &
\subfloat{\includegraphics[width=0.19\textwidth]{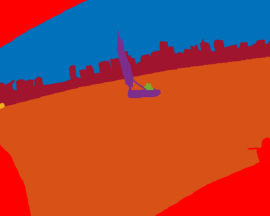}} &
\subfloat{\includegraphics[width=0.19\textwidth]{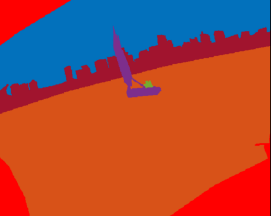}} &   
\subfloat{\includegraphics[width=0.19\textwidth]{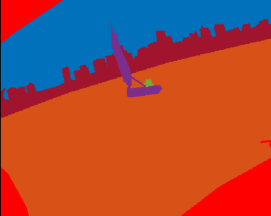}} &
\subfloat{\includegraphics[width=0.19\textwidth]{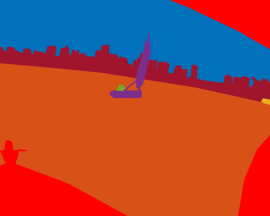}} 
\end{tabular}
\caption{Data augmentation: First column shows original image and its mask followed by 3 rotations: 2, 5 and 7 degrees respectively. Last column shows mirror operation. Images were also rotated by -2, -5 and -7 degrees.}
\label{data_augment}
\end{figure}

Robust, diverse and large volume of data is a prerequisite for any successful model. However, creating a labeled dataset is a laborious and expensive task. A common technique employed in increasing the dataset variety is data augmentation \cite{scholler}, \cite{mastr}. We used the following 2 types of transformations for our dataset: 
\begin{itemize}
    \item Rotation: An image and its mask were rotated by $\pm 2$\textdegree, $\pm 5$\textdegree, $\pm 7$\textdegree
    \item Mirror: The original image and mask and their rotated equivalence from previous step was mirrored on the vertical axis
\end{itemize}
This scheme generated 13 images from 1 image. Accordingly, our dataset was augmented to 40,096 images. An illustrative example is shown in Fig. \ref{data_augment}. 

\subsection{Performance Criterion}
Intersection over union (IoU) is an important metric for segmentation inference \cite{csurka2013good}. It is defined as ratio of area of intersection between predicted segmentation mask (P) and the ground truth (G), and the area of union between the two.
\vspace{1ex}
\begin{equation}
\footnotesize
\mathrm{IoU = \dfrac{G\cap P}{G\cup P}}
\end{equation}
\vspace{1ex}
The IoU score ranges between 0 and 1. A threshold value can be chosen to determine if the inference can be treated as true positive (TP). If the IoU score is greater than a certain threshold then we treat the inference as true positive (TP) otherwise it is treated as false negative (FN). If the ground truth does not contain the region present in prediction then it is treated as false positive (FP). Usually a threshold value of greater than 0.5 is considered optimal \cite{iou_threshold}, however, if the class sizes are tiny as in the present case, the threshold can be compromised to a lower value. The precision, recall and F1 score are defined as follows:
\begin{equation}
\footnotesize
Precision(Pr) = \dfrac{TP}{TP + FP}, \enspace Recall(Re) = \dfrac{TP}{TP + FN} 
\end{equation}
\begin{equation}
\footnotesize
    F1 = \dfrac{2 * Precision * Recall}{Precision + Recall}  
\end{equation}

\subsection{Performance Metrics}
All 3 architectures were run on \cite{google_colab} platform which provided a 16GB GPU memory. Training was performed from scratch without using pretrained weights. Images were randomly shuffled prior to the training which helped in improving the robustness of the model. Table \ref{performance} gives a summary of the high level metrics obtained for each architecture.

\begin{figure*}[th] 
\setlength{\tabcolsep}{1pt}
\centering
\begin{tabular}{cccc}
\scriptsize \textbf{Test}  & \scriptsize \textbf{UNet} & \scriptsize \textbf{PSPNet} & \scriptsize \textbf{DeepLabv3} \\
\subfloat{\includegraphics[width=0.23\textwidth]{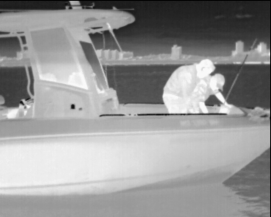}}  &  
\subfloat{\includegraphics[width=0.23\textwidth]{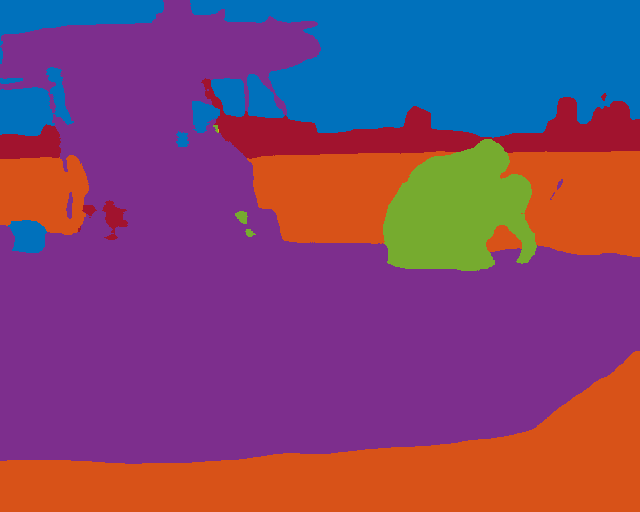}}  &  
\subfloat{\includegraphics[width=0.23\textwidth]{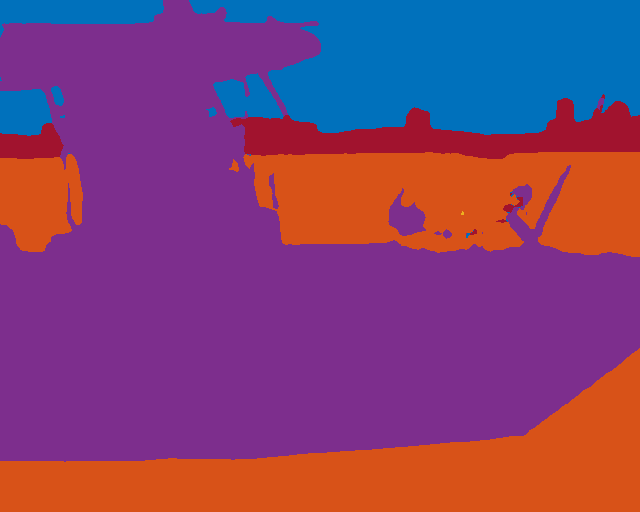}}  &  
\subfloat{\includegraphics[width=0.23\textwidth]{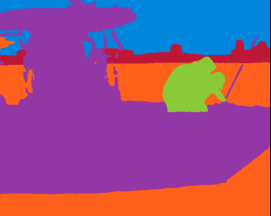}}   
\\
\subfloat{\includegraphics[width=0.23\textwidth]{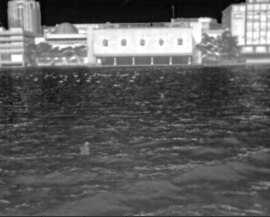}} & 
\subfloat{\includegraphics[width=0.23\textwidth]{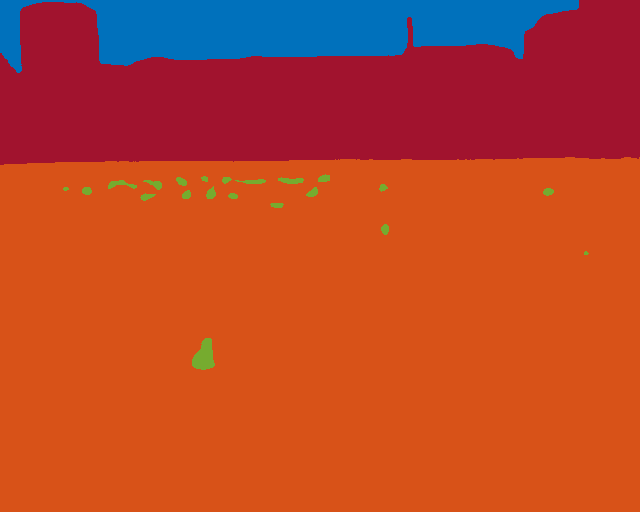}} & 
\subfloat{\includegraphics[width=0.23\textwidth]{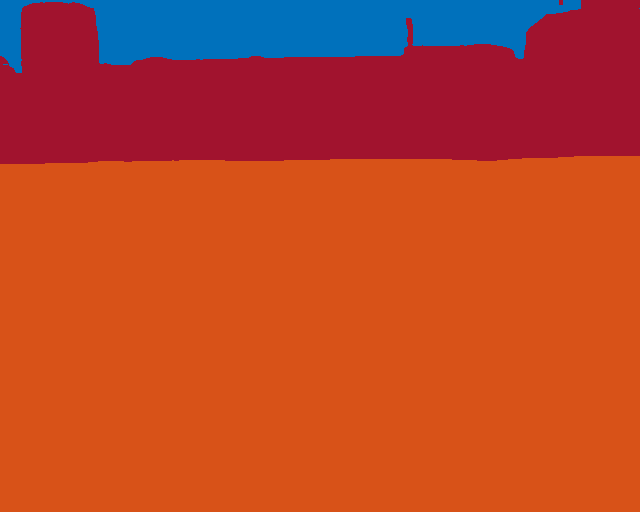}} & 
\subfloat{\includegraphics[width=0.23\textwidth]{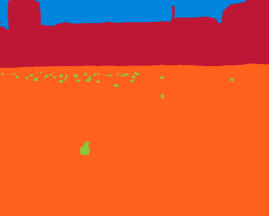}} 
\\
\subfloat{\includegraphics[width=0.23\textwidth]{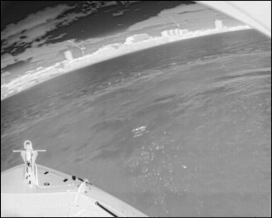}} &  
\subfloat{\includegraphics[width=0.23\textwidth]{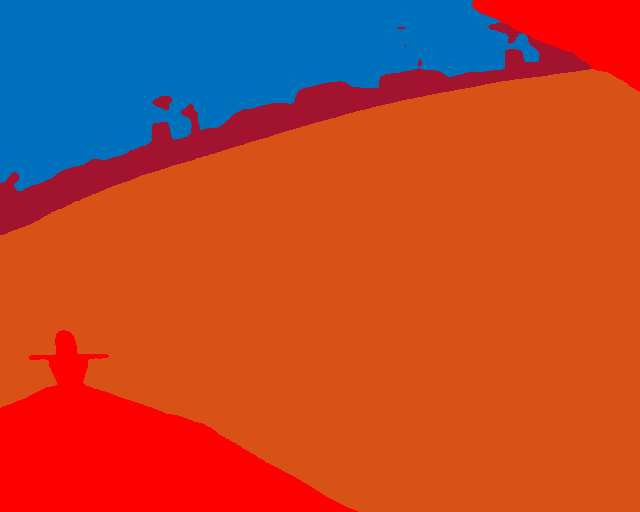}} &  
\subfloat{\includegraphics[width=0.23\textwidth]{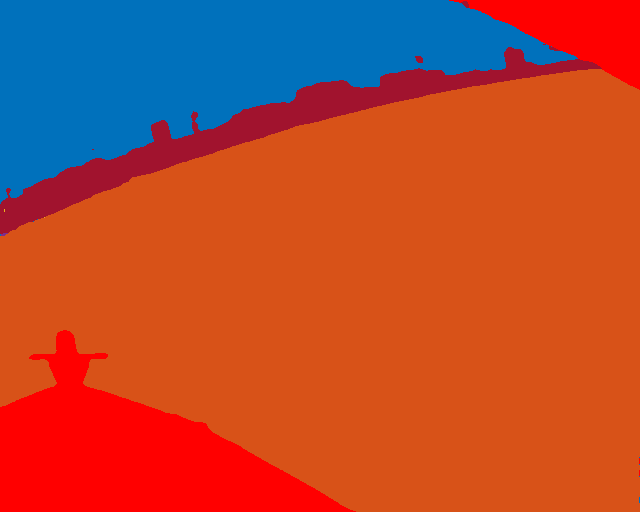}} &  
\subfloat{\includegraphics[width=0.23\textwidth]{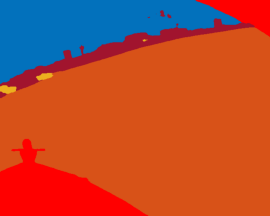}}  
\\
\subfloat{\includegraphics[width=0.23\textwidth]{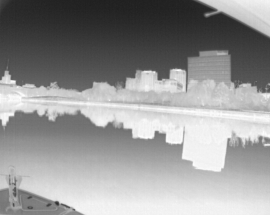}} &  
\subfloat{\includegraphics[width=0.23\textwidth]{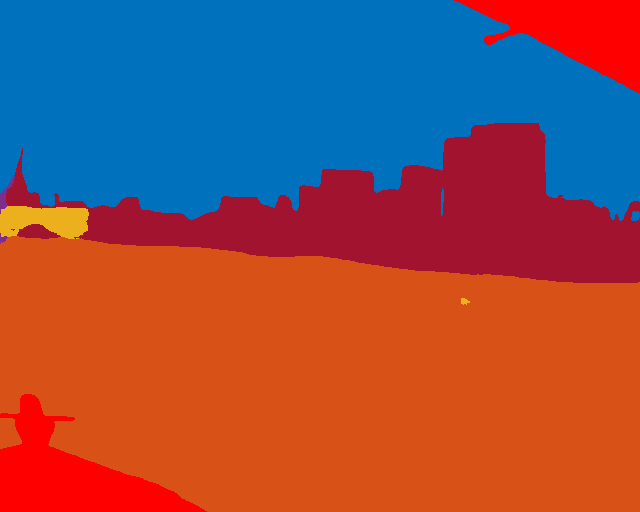}} &  
\subfloat{\includegraphics[width=0.23\textwidth]{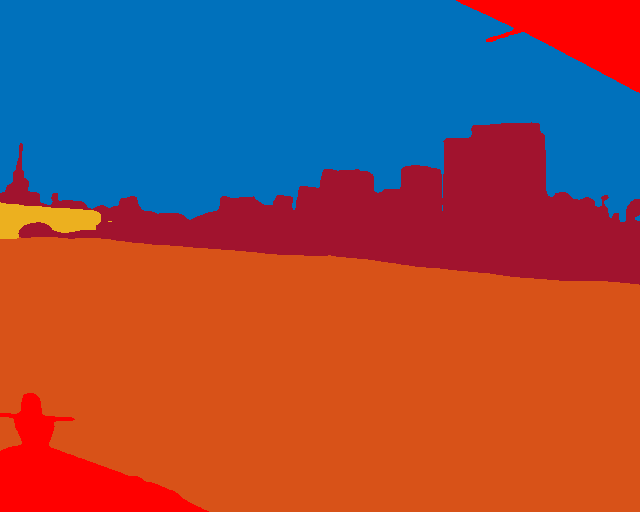}} &  
\subfloat{\includegraphics[width=0.23\textwidth]{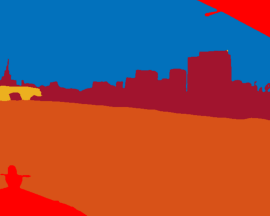}}  
\\
\subfloat{\includegraphics[width=0.23\textwidth]{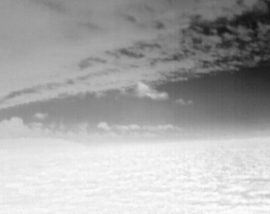}} & 
\subfloat{\includegraphics[width=0.23\textwidth]{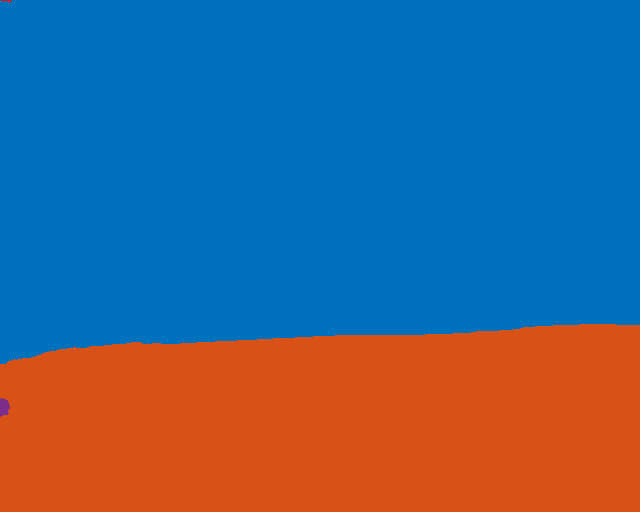}} &  
\subfloat{\includegraphics[width=0.23\textwidth]{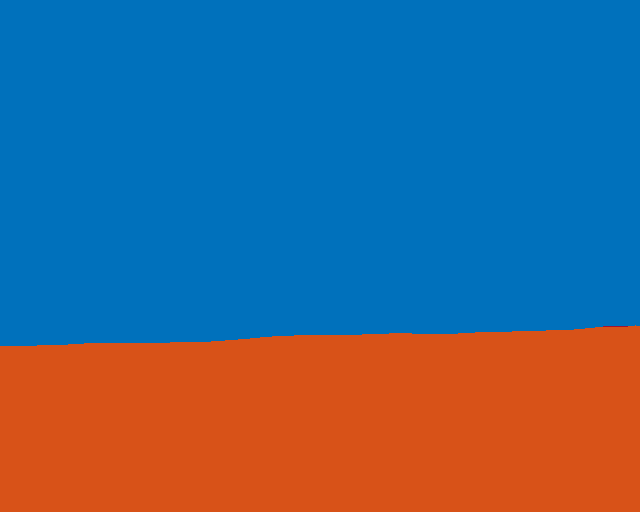}} &  
\subfloat{\includegraphics[width=0.23\textwidth]{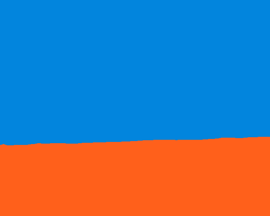}}   
\end{tabular}
\caption{Qualitative comparison of the inference: First row: DeepLabv3 and UNet have a good inference other than a few traces of sky in the obstacle. PSPNet, despite the significant size has problem in identifying living obstacles and will need significant additional training. Second row: Very tiny birds in the water have been correctly captured by both DeepLabv3 and UNet and are again missed in PSPNet. Third row: Curvature of the image and waves in water have been correctly depicted in all 3 classifiers. DeepLabv3 does a good job in calling out the bridges as well. Fourth row: There is significant reflection in the water that is handled well by all, except for a few sporadic traces of bridge. Fifth row: There is a very obscure cloud and sky boundary. This has been handled really well in all 3 models. Overall, glitter, waves, horizon boundaries that are a problem in optical world are handled well in the LWIR dataset. DeepLabv3 performs best overall, closely followed by PSPNet (except for the problem with living obstacle) and UNet}
\label{inference_correct}
\end{figure*}

\begin{table}[th]
\caption{Performance of various architectures: Size of the images and number of epochs were identical across the 3 architectures.}
\label{performance}
\scriptsize
\begin{center}
\resizebox{1.0\columnwidth}{!}{%
\begin{tabular}{ m{10mm}  m{10mm}  m{7mm}  m{5mm} m{8mm} m{12mm} }
\hline
\rowcolor{lightgray}\textbf{Name}  & \textbf{Images} & \textbf{Image size} & \textbf{Epoch} & \textbf{Loss} & \textbf{Training time}   \\
\hline
UNet &\multirow{3}{7mm}{train:27678 val:7910 test:4508} &\multirow{3}{*}{640*512} &\multirow{3}{*}{50} & 0.0202 & 27hr 50min  \\
PSPNet &  &  &  &  0.0223 & 47hr 56min \\
DeepLabv3 &  &  &  &  0.0112 & 29hr 30min \\
\hline
\end{tabular}%
}
\end{center}
\end{table}

\begin{table}[th]
\caption{Quantitative results on the MassMIND dataset: Classes pertinent to the navigation path were also evaluated against a lower threshold. PSPNet performs at par with DeepLabv3 in all classes except for 'living obstacles' as indicated by the poor recall. Most of the times it identifies the living obstacle but collapses it in the obstacle (a boat typically). Large number of epochs for PSPNet will probably circumvent this problem. F1 score is really good for sky, water, background and self class as expected. Despite their tiny size, obstacles give a satisfactory recall. DeepLabv3 has performed well across all the classes.}   
\label{evaluation_results}
\footnotesize
\begin{center}

\resizebox{1.0\columnwidth}{!}{%
\begin{tabular}{llllllllll}
\hline
\rowcolor{lightgray}\textbf{Model} & \textbf{Class} & \textbf{Th} & \textbf{TP} & \textbf{FP} & \textbf{FN} & \textbf{Pr} & \textbf{Re} & \textbf{F1}  \\
\hline
UNet & \multirow{3}{*}{Sky} & \multirow{3}{*}{0.6} & 4327 & 24 & 153 & 99.4 & 96.6 & 98.0  \\
DeepLabv3 &  &  & 4374 & 8 & 106 & 99.8 & 97.6 & 98.7  \\
PSPNet &  &  & 4408 & 18 & 72 & 99.6 & 98.4 & 99.0  \\
\hline 

UNet & \multirow{3}{*}{Water}  & \multirow{3}{*}{0.6} & 4508 & 0 & 0 & 100 & 100 & 100 \\
DeepLabv3 &  &  & 4508 & 0 & 0 & 100 & 100 & 100 \\
PSPNet &  &  & 4508 & 0 & 0 & 100 & 100 & 100 \\
\hline 

\multirow{ 2}{*}{UNet} &\multirow{6}{*}{Bridge} & 0.6 & 723 & 605 & 409 & 54.4 & 63.9 & 58.8 \\ 
& &  0.3 & 962 & 605 & 170 & 61.4 & 85.0 & 71.3 \\ \cline{4-9}
\multirow{2}{*}{DeepLabv3} &  & 0.6 & 870 & 377 & 262 & 69.8 & 76.9 & 73.1 \\ 
& &  0.3 & 1013 & 377 & 119 & 72.9 & 89.5 & 80.3 \\ \cline{4-9}
\multirow{2}{*}{PSPNet} &  & 0.6 & 910 & 319 & 222 & 74.0 & 80.4 & 77.1 \\ 
& &  0.3 & 1040 & 319 & 92 & 76.5 & 91.9 & 83.5 \\ 
\hline 

\multirow{2}{*}{UNet} &\multirow{6}{*}{Obstacle} & 0.6 & 615 & 504 & 2193 & 55.0 & 21.9 & 31.3 \\ 
& &  0.3 & 1645 & 504 & 1163 & 76.5 & 58.6 & 66.4 \\ \cline{4-9}
\multirow{2}{*}{DeepLabv3} &  & 0.6 & 1143 & 429 & 1665 & 72.7 & 40.7 & 52.2  \\ 
& &  0.3 & 2045 & 429 & 763 & 82.7 & 72.8 & 77.4 \\ \cline{4-9}
\multirow{2}{*}{PSPNet} &   & 0.6 & 1175 & 428 & 1633 & 73.3 & 41.8 & 53.3 \\ 
& &  0.3 & 2068 & 428 & 740 & 82.9 & 73.6 & 78.0 \\ 
\hline 

\multirow{2}{*}{UNet} &\multirow{6}{*}{Living Ob}  & 0.6 & 98 & 136 & 886 & 41.9 & 10.0 & 16.1 \\ 
& &  0.3 & 420 & 136 & 564 & 75.5 & 42.7 & 54.5 \\ \cline{4-9}
\multirow{2}{*}{DeepLabv3} &  & 0.6 & 294 & 141 & 690 & 67.6 & 29.9 & 41.4  \\ 
& &  0.3 & 599 & 141 & 385 & 80.9 & 60.9 & 69.5 \\ \cline{4-9}
\multirow{2}{*}{PSPNet} &  & 0.6 & 65 & 9 & 1045 & 87.8 & 5.9 & 11.0 \\ 
& &  0.3 & 127 & 9 & 983 & 93.4 & 11.4 & 20.4 \\
\hline

UNet & \multirow{3}{*}{Background}  & \multirow{3}{*}{0.6} & 3856 & 85 & 526 & 97.8 & 88.0 & 92.7  \\
DeepLabv3 &  &  & 4049 & 33 & 333 & 99.2 & 92.4 & 95.7 \\
PSPNet &  &  & 4178 & 50 & 204 & 98.8 & 95.3 & 97.0 \\
\hline 

UNet & \multirow{3}{*}{Self}  & \multirow{3}{*}{0.6} & 2306 & 179 & 74 & 92.8 & 96.9 & 94.8 \\
DeepLabv3 &   &  & 2322 & 54 & 58 & 97.7 & 97.6 & 97.6 \\
PSPNet &   &  & 2329 & 19 & 51 & 99.2 & 97.9 & 98.5 \\
\hline 
\end{tabular}%
}
\end{center}
\end{table}

The training loss is quite low for all the three architectures. However, since water and sky occupy significant portions of the maritime images, this parameter does not help much and gives misleading accuracy. Classes were therefore individually analyzed and Table \ref{evaluation_results} reports these metrics at a class level. As noted above, sky and water have a significant contribution in each image and therefore yield a high F1 even with a threshold of 0.6. The classes that are critical for path navigation namely `obstacles', `living obstacles' and `bridge' are much smaller as per Fig. \ref{instance_concentration}. As shown in Table \ref{evaluation_results}, a threshold of 0.6 gives substantial FNs for `'obstacles' and `living obstacles'. Since the pixel area of these classes is really tiny, we decided to evaluate their performance by lowering the threshold to 0.3. As seen from the table, the FN reduced appreciably. The threshold value can be subjective based on the size of the class. A value of 0.6 for tiny sizes is too high a bar to achieve. The primary intent of the model is path navigation and having a dynamic threshold based on parameters like size or distance would enable the model to predict the presence of an obstacle although it may have lesser spatial accuracy. Poor recall score of PSPNet on `living obstacles' indicates its failure to identify them which is also reflected in low F1 score.
\par Fig. \ref{inference_correct} indicates the inference results from all 3 architectures for a few sample images. Living obstacles (humans) have been inferred really well in both DeepLabv3 and UNet. The boundaries separating humans from the obstacle (boat) has been interpreted well. In the second row, very tiny obstacles (birds) in the water have been correctly inferred by DeepLabv3 and UNet. Third row indicates that curvature or waves in water pose no issue to the model. Reflection in water in the fourth row, or the heavy cloud cover right on top of the sea, has been interpreted properly by the models. PSPNet was found to have problems with identifying `living obstacles'. While it works really well for all other classes, it needs a lot of training to consistently differentiate the boundaries between obstacle and the living obstacle. \par
\begin{figure*}[th] 
\setlength{\tabcolsep}{1pt}
\begin{tabular}{cccc}
\scriptsize \textbf{Test}  & \scriptsize \textbf{UNet} & \scriptsize \textbf{PSPNet} & \scriptsize \textbf{DeepLabv3} \\
\subfloat{\includegraphics[width=0.23\textwidth]{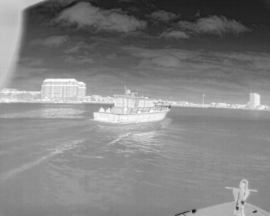}}  &  
\subfloat{\includegraphics[width=0.23\textwidth]{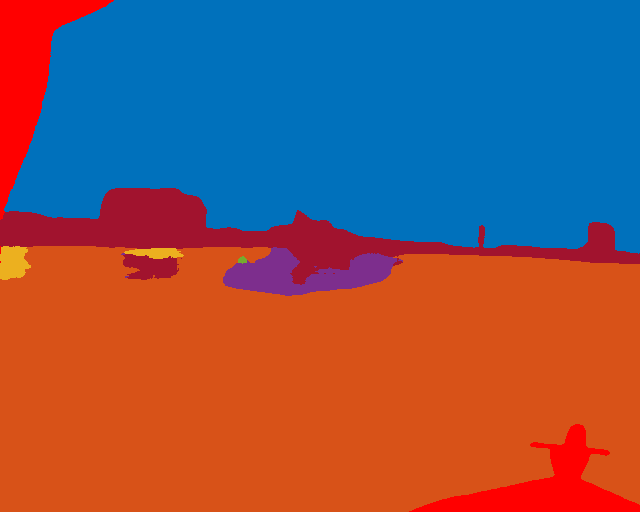}}  &  
\subfloat{\includegraphics[width=0.23\textwidth]{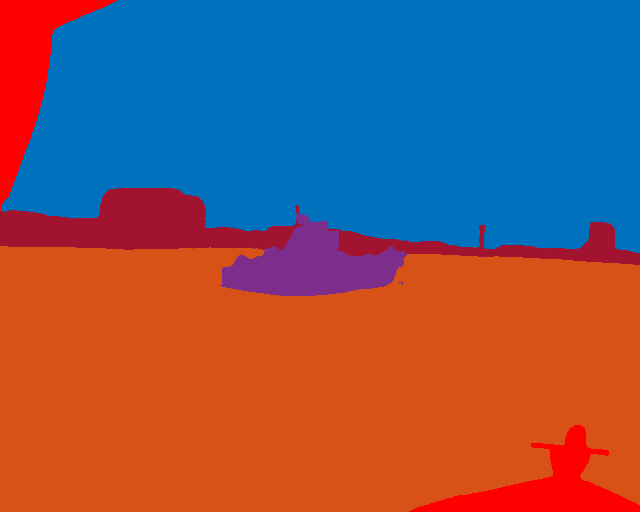}}  &  
\subfloat{\includegraphics[width=0.23\textwidth]{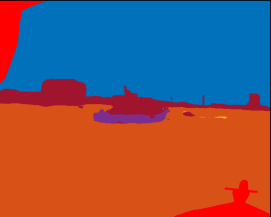}}  
\\
\subfloat{\includegraphics[width=0.23\textwidth]{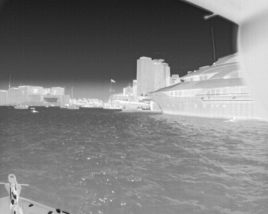}}  &  
\subfloat{\includegraphics[width=0.23\textwidth]{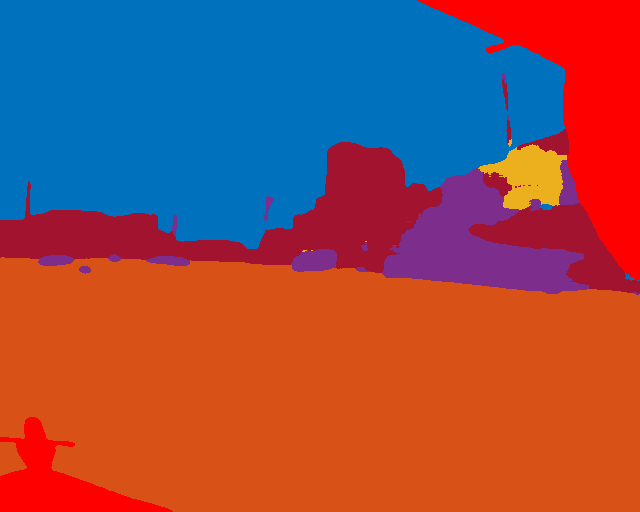}}  &  
\subfloat{\includegraphics[width=0.23\textwidth]{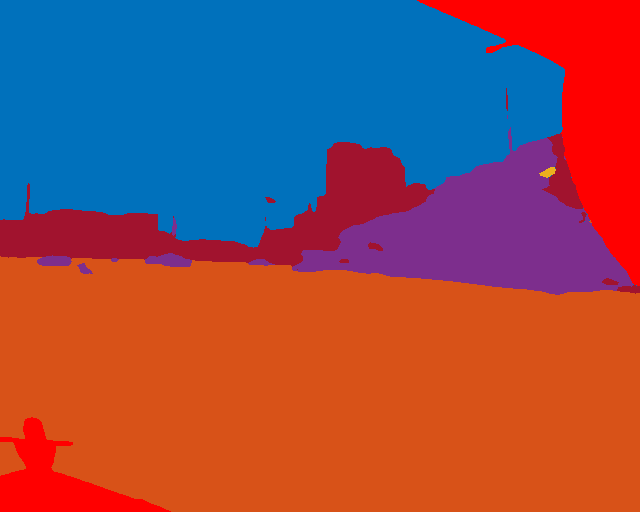}}  &  
\subfloat{\includegraphics[width=0.23\textwidth]{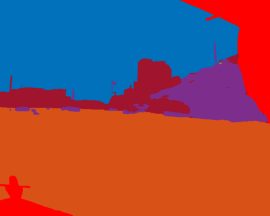}}  
\\
\subfloat{\includegraphics[width=0.23\textwidth]{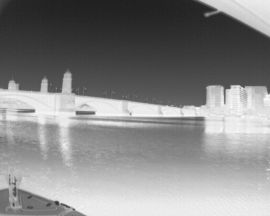}}  &  
\subfloat{\includegraphics[width=0.23\textwidth]{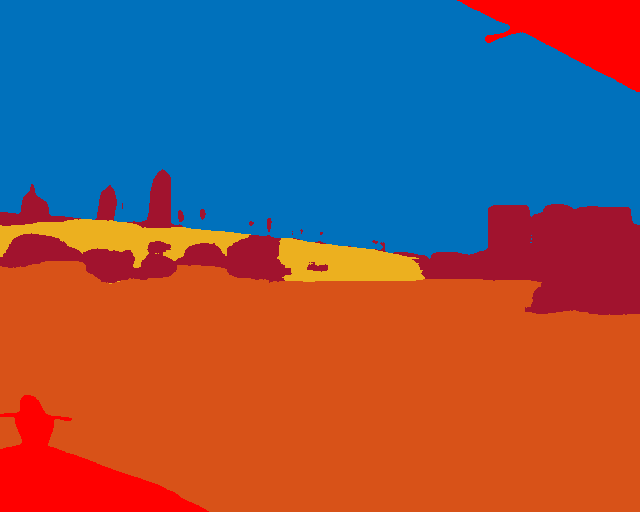}}  &  
\subfloat{\includegraphics[width=0.23\textwidth]{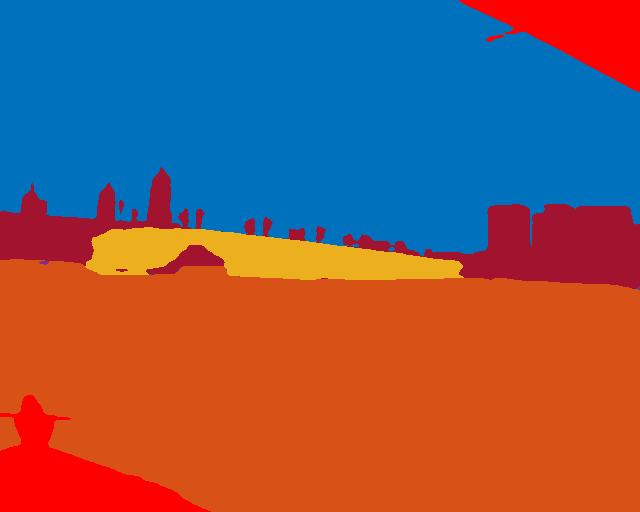}}  &  
\subfloat{\includegraphics[width=0.23\textwidth]{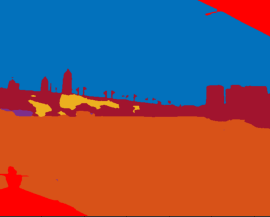}}  
\end{tabular}
\caption{Misrepresented inference: First row: A portion of the obstacle has been clubbed with the background in both UNet and DeepLabv3, however it is represented properly in PSPNet. There are sporadic traces of bridge in UNet. Second row:  Again, the hull of the larger ship is very close to the background and the classifier has problems detecting the boundaries in all 3 architectures. And there are a few spurious occurrences of bridge and living obstacles. 
Third row: PSPNet has a better inference than the rest although the boundaries are not clear. Portions of bridge and background have been collapsed in the other two along with a few sporadic obstacles. The reflection in water has been handled well in all three.}
\label{inference_incorrect}
\end{figure*}

 In addition to the above limitation in PSPNet, there were some situations where very tiny obstacles were missed or not completely identified. Bridge is another area where the classifiers sometimes gave inconsistent results. This was primarily due to the fact that ground truth for the bridge contained portion annotated as `background' that is not pertinent to navigation. The classifier therefore sometimes incorrectly inferred the navigable portion of the bridge as background and vice versa. Despite that, inference for `bridge' class improved significantly after increasing the training to 50 epochs. Some of the challenge cases are shown in Fig. \ref{inference_incorrect}.    

To get consistent and meaningful performance from the deep learning models, various modifications were tried and parameters changed. A few of those key modifications are elaborated below: 
\par Initially we had reduced the image size to 320$\times$256(width$\times$height) to speedup the training phase. This resulted in a lot of smaller obstacles being missed in the inference. The obstacles were inherently small as shown in Table \ref{size_variation} and they got tinier with the reduction in image size leaving the classifier with very less information to train well. It was therefore decided to retain the original size of the images. We also experimented with excluding 2019 images because they were of low resolution. We expected improvement in the performance. However,the performance worsened as a lot of variation and information contained in the 2019 dataset was lost. We therefore retained both 2019 and 2020 images.    
\par To increase the dataset size, data augmentation was used as described in Sub-Section 5.1. For optical images, color adjustment is yet another augmentation operation. However, augmentation by changing the brightness of LWIR images had adverse effect on the inference and was not incorporated. Larger rotations ($\pm 10$\textdegree, $\pm 15$\textdegree) were tried and that caused distortion in 2019 images and deteriorated the overall results. Smaller range ($\pm 2$\textdegree, $\pm 5$\textdegree, $\pm 7$\textdegree) seem to fit the dataset well. An important consideration that improved the performance drastically was shuffling. Since our images were named based on epoch times, without explicit shuffling they were fed sequentially to the classifier and gave poor results. They were shuffled randomly before the training and that variation helped the classifiers boost the performance multi-fold. 
\par As seen from the inference, horizon detection does not pose a problem in LWIR images. However, there are some sporadic instances of clouds being incorrectly inferred as water and vice versa. This happens mainly when there are white clouds in the sky that appear similar to water, have similar pixel values and get labeled incorrectly. Bridge annotation remained challenging despite good F1 scores. Depending on the nature of the bridge and position of the background with respect to the bridge, the classifier has a challenge and classifies the navigable portion of the bridge as a background. When the bridge has arches, view from beneath the arch could be sky or background and water. Such minute details pleasantly have been represented well most of the times, but do pose a challenge in other cases.  
\par In summary, standard CNN architectures can be trained with LWIR images and can produce good inference. The LWIR dataset indeed complements the optical dataset in many such scenarios and is therefore invaluable in the perception task.

\section {CONCLUSION}
We have presented a comprehensive MassMIND dataset containing over 2,900 segmented diverse LWIR images and made it publicly available. This is the first maritime dataset that segments LWIR images in 7 different classes with emphasis on living and non-living, static and dynamic obstacles in the water. The dataset is also evaluated and bench marked against a few industry standard architectures.  
\par Of the 3 classifiers evaluated, DeepLabv3 \cite{deeplab} stands out the best, followed by UNet and PSPNet. PSPNet is found to be comparable to DeepLabv3 in all other classes except for living obstacles. Our research calls out significance of LWIR when it comes to problems such as  extreme light conditions and horizon detection that are common in the optical world. In addition, the dataset tries to encompass many of the critical aspects of the maritime imagery such as static and dynamic obstacles of varying sizes, crowded shoreline, deep seas, varying weather and light patterns, and life presence on the water surface. Such a diverse LWIR dataset can be useful to improve the deep learning architectures for the maritime environment. At the same time, it can complement optical datasets.
The present dataset also has some limitations. Certain obstacles such as river barges or Sea-Doos are not captured because of the ASVs' urban setting. The demographics within the dataset is also limited to one region in USA. We understand that as a result, there is no variety in navigational markers used across world regions. This gap can be bridged by collaboration with other researchers.
\par We plan to refine the current 'obstacle' class to indicate the type of obstacle such as sailboats, kayaks, and merchant ships. This will enable creation of COLREG compliant classifiers. We plan to integrate the inference obtained from these models in real time with the perception pipeline developed by \cite{tom_perception}. Lastly, we will also explore creation of an instance-segmentation architecture dedicated  towards LWIR images in the marine environment.

\begin{acks}
We are grateful to Brunswick\textsuperscript{\textregistered} Corporation for their support. The views, opinions and/or findings expressed are those of the authors and should not be interpreted as representing the official views of Brunswick\textsuperscript{\textregistered} Corporation.
We also thank the team at segments.ai for providing early access free of charge and for providing dashboard for tracking the progress of the labeling work.
\end{acks}

The Authors declare that there is no conflict of interest.

\bibliographystyle{Sageh}
\bibliography{dataset.bib}

\begin{thebibliography}{36}
\providecommand{\natexlab}[1]{#1}
\providecommand{\url}[1]{\texttt{#1}}
\providecommand{\urlprefix}{URL }
\expandafter\ifx\csname urlstyle\endcsname\relax
  \providecommand{\doi}[1]{DOI:\discretionary{}{}{}#1}\else
  \providecommand{\doi}{DOI:\discretionary{}{}{}\begingroup
  \urlstyle{rm}\Url}\fi

\bibitem[{AUVLab(1989)}]{auv}
AUVLab (1989) {MIT Sea Grant, Autonomous Underwater Vehicles Lab}
  \urlprefix\url{https://seagrant.mit.edu/auv-lab/}.

\bibitem[{Bloisi et~al.(2015)Bloisi, Iocchi, Pennisi and Tombolini}]{mardct}
Bloisi D, Iocchi L, Pennisi A and Tombolini L (2015) {ARGOS-Venice Boat
  Classification}.
\newblock \emph{12th IEEE International Conference on Advanced Video and Signal
  Based Surveillance (AVSS)} : 1--6\doi{10.1109/AVSS.2015.7301727}.

\bibitem[{Bovcon and Kristan(2020)}]{wasr}
Bovcon B and Kristan M (2020) A water-obstacle separation and refinement
  network for unmanned surface vehicles.
\newblock \emph{2020 {IEEE} International Conference on Robotics and
  Automation, {ICRA} 2020, Paris, France, May 31 - August 31, 2020} :
  9470--9476\doi{10.1109/ICRA40945.2020.9197194}.
\newblock \urlprefix\url{https://doi.org/10.1109/ICRA40945.2020.9197194}.

\bibitem[{Bovcon et~al.(2019)Bovcon, Muhovi\v{c}, Per\v{s} and Kristan}]{mastr}
Bovcon B, Muhovi\v{c} J, Per\v{s} J and Kristan M (2019) The {MaSTr1325}
  dataset for training deep usv obstacle detection models.
\newblock \emph{IEEE/RSJ International Conference on Intelligent Robots and
  Systems (IROS)} .

\bibitem[{Bovcon et~al.(2021)Bovcon, Muhovi\v{c}, Vranac, Mozeti\v{c}, Per\v{s}
  and Kristan}]{mods}
Bovcon B, Muhovi\v{c} J, Vranac D, Mozeti\v{c} D, Per\v{s} J and Kristan M
  (2021) {MODS- A USV-oriented object detection and obstacle segmentation
  benchmark} .

\bibitem[{BSD(2020)}]{rosbag}
BSD (2020) Rosbag package \urlprefix\url{http://wiki.ros.org/rosbag}.

\bibitem[{Castellini et~al.(2020)Castellini, Bloisi, Blum, Masillo and
  Farinelli}]{intcatch}
Castellini A, Bloisi D, Blum J, Masillo F and Farinelli A (2020) {Intcatch
  Aquatic Drone Sensor Dataset}.
\newblock \emph{Mendeley Data, V2} \doi{10.17632/gtt7stf5x8.2}.

\bibitem[{Chen et~al.(2018)Chen, Papandreou, Kokkinos, Murphy and
  Yuille}]{deeplab}
Chen L, Papandreou G, Kokkinos I, Murphy K and Yuille A (2018) {Deeplab:
  Semantic image segmentation with deep convolutional nets, atrous convolution,
  and fully connected CRFs}.
\newblock \emph{IEEE TPAMI} 40(4): 834--848.

\bibitem[{Clunie et~al.(2021)Clunie, DeFilippo, Sacarny and
  Robinette}]{tom_perception}
Clunie T, DeFilippo M, Sacarny M and Robinette P (2021) Development of a
  perception system for an autonomous surface vehicle using monocular camera,
  lidar, and marine radar.
\newblock \emph{International Conference on Robotics and Automation} .

\bibitem[{Colab(2021)}]{google_colab}
Colab (2021) Making the most of your {C}olab
  \urlprefix\url{https://colab.research.google.com/}.

\bibitem[{Cordts et~al.(2016)Cordts, Omran, Ramos, Rehfeld, Enzweiler,
  Benenson, Franke, Roth and Schiele}]{cityscapes}
Cordts M, Omran M, Ramos S, Rehfeld T, Enzweiler M, Benenson R, Franke U, Roth
  S and Schiele B (2016) {The Cityscapes Dataset for Semantic Urban Scene
  Understanding}.
\newblock \emph{Proc. of the IEEE Conference on Computer Vision and Pattern
  Recognition (CVPR)} .

\bibitem[{Csurka et~al.(2013)Csurka, Larlus, Perronnin and
  Meylan}]{csurka2013good}
Csurka G, Larlus D, Perronnin F and Meylan F (2013) What is a good evaluation
  measure for semantic segmentation?.
\newblock \emph{Bmvc} 27(2013): 10--5244.

\bibitem[{Debals(2021)}]{segments}
Debals O (2021) Platform for {I}mage {S}egmentation
  \urlprefix\url{https://segments.ai}.

\bibitem[{DeFilippo et~al.(2021)DeFilippo, Sacarny and Robinette}]{robowhaler}
DeFilippo M, Sacarny M and Robinette P (2021) Robo{W}haler: A {R}obotic
  {V}essel for {M}arine {A}utonomy and {D}ataset {C}ollection.
\newblock \emph{OCEANS} .

\bibitem[{Deng et~al.(2009)Deng, Dong, Socher, Li, Li and Fei-Fei}]{imagenet}
Deng J, Dong W, Socher R, Li LJ, Li K and Fei-Fei L (2009) Imagenet: A large
  scale hierarchical database.
\newblock \emph{CVPR} : 248--255.

\bibitem[{EO(2022)}]{edmund_optics}
EO (2022) What is {SWIR}?
  \urlprefix\url{https://www.edmundoptics.com/knowledge-center/application-notes/imaging/what-is-swir/?utm_medium=email&utm_source=transaction}.

\bibitem[{Hyll(2016)}]{nir_vs_lwir_compare}
Hyll K (2016) Image-based quantitative infrared analysis and microparticle
  characterisation for pulp and paper applications
  \doi{10.13140/RG.2.1.2595.6884}.

\bibitem[{IMO(1972)}]{colreg}
IMO (1972) {COLREGs: Convention on the International Regulations for Preventing
  Collisions at Sea} .

\bibitem[{Krizhevsky et~al.(2012)Krizhevsky, Sutskever and Hinton}]{cnn}
Krizhevsky A, Sutskever I and Hinton G (2012) Imagenet classification with deep
  convolutional neural networks.
\newblock \emph{Advances in Neural Information Processing Systems} 25:
  1097--1105.

\bibitem[{Liu et~al.(2021)Liu, Geng, Zhao, Zhang and Li}]{iou_threshold}
Liu J, Geng Y, Zhao J, Zhang K and Li W (2021) {Image Semantic Segmentation Use
  Multiple-threshold Probabilistic R-CNN with Feature Fusion}.
\newblock \emph{Symmetry} .

\bibitem[{NASA(2003)}]{nasa_thermal}
NASA (2003) Near, {M}id and {F}ar {I}nfrared
  \urlprefix\url{http://www.icc.dur.ac.uk/~tt/Lectures/Galaxies/Images/Infrared/Regions/irregions.html}.

\bibitem[{Nirgudkar et~al.(2022)Nirgudkar, DeFilippo, Sacarny, Benjamin and
  Robinette}]{our_dataset}
Nirgudkar S, DeFilippo M, Sacarny M, Benjamin M and Robinette P (2022)
  {MassMIND: Massachusetts Maritime INfrared Dataset}
  \urlprefix\url{https://github.com/uml-marine-robotics/MassMIND}.

\bibitem[{Nirgudkar and Robinette(2021)}]{shailesh_icar2021}
Nirgudkar S and Robinette P (2021) {Beyond visible light: Object Detection
  Using Long Wave Infrared Images in Maritime Environment}.
\newblock \emph{International Conference on Autonomous Robotics} .

\bibitem[{Patino et~al.(2016)Patino, Cane, Vallee and Ferryman}]{ipatch}
Patino L, Cane T, Vallee A and Ferryman J (2016) {PETS 2016: Dataset and
  Challenge}.
\newblock \emph{IEEE Conference on Computer Vision and Pattern Recognition
  Workshops (CVPRW)} : 1240--1247\doi{10.1109/CVPRW.2016.157}.

\bibitem[{Prasad et~al.(2017)Prasad, Rajan, Rachmawati, Rajabaly and
  Quek}]{smd}
Prasad DK, Rajan D, Rachmawati L, Rajabaly E and Quek C (2017) {Video
  Processing from Electro-optical Sensors for Object Detection and Tracking in
  Maritime Environment: A Survey}.
\newblock \emph{IEEE Transactions on Intelligent Transportation Systems} 18(8):
  1993--2016.

\bibitem[{Qi et~al.(2011)Qi, Wu, Dai and He}]{qi}
Qi B, Wu T, Dai B and He H (2011) Fast detection of small infrared objects in
  maritime scenes using local minimum patterns.
\newblock \emph{18th IEEE International Conference on Image Processing} :
  3553--3556\doi{10.1109/ICIP.2011.6116483}.

\bibitem[{Ribeiro(2015)}]{seagull}
Ribeiro R (2015) {The Sea-gull Dataset}
  \urlprefix\url{http://vislab.isr.ist.utl.pt/seagull-dataset}.

\bibitem[{Robinette et~al.(2019)Robinette, Sacarny, DeFilippo, Novitzky and
  Benjamin}]{sensor_evaluation_robinette}
Robinette P, Sacarny M, DeFilippo M, Novitzky M and Benjamin MR (2019) Sensor
  evaluation for autonomous surface vehicles in inland waterways.
\newblock \emph{OCEANS} .

\bibitem[{Rodin and Johansen(2018)}]{rodin}
Rodin CD and Johansen TA (2018) {Detectability of Objects at the Sea Surface in
  Visible Light and Thermal Camera Images}.
\newblock \emph{OCEANS - MTS/IEEE Kobe Techno-Oceans (OTO)}
  \doi{10.1109/OCEANSKOBE.2018.8559310}.

\bibitem[{Ronneberger et~al.(2015)Ronneberger, Fischer and Brox}]{unet}
Ronneberger O, Fischer P and Brox T (2015) {U-Net: Convolutional Networks for
  Biomedical Image Segmentation}.
\newblock \emph{Medical Image Computing and Computer-Assisted Intervention
  (MICCAI)} 9351: 234--241.
\newblock \doi{10.1007/978-3-319-24574-4_28}.

\bibitem[{Sch\"{o}ller et~al.(2019)Sch\"{o}ller, Plenge-Feidenhans'l, Stets and
  Blanke}]{scholler}
Sch\"{o}ller FET, Plenge-Feidenhans'l MK, Stets JD and Blanke M (2019)
  {Assessing Deep-learning Methods for Object Detection at Sea from {LWIR}
  Images}.
\newblock \emph{International federation of automatic control workshop} 52(21):
  64--71.
\newblock \doi{10.1109/OCEANSKOBE.2018.8559310}.

\bibitem[{SeaGrant(2022)}]{unannotated_dataset}
SeaGrant (2022) {MIT Sea Grant Marine Perception Dataset}
  \urlprefix\url{https://seagrant.mit.edu/auvlab-datasets-marine-perception-1}.

\bibitem[{Steccanella et~al.(2020)Steccanella, Bloisi, Castellini and
  Farinelli}]{intcatch2}
Steccanella L, Bloisi D, Castellini A and Farinelli A (2020) {Waterline and
  Obstacle Detection in Images from Low-Cost Autonomous Boats for Environmental
  Monitoring}.
\newblock \emph{Robotics and Autonomous Systems}
  \doi{10.1016/j.robot.2019.103346}.

\bibitem[{Wang et~al.(2017)Wang, Dong, Zhao and Xu}]{wang}
Wang B, Dong L, Zhao M and Xu W (2017) Fast infrared maritime target detection:
  Binarization via histogram curve transformation.
\newblock \emph{Infrared Physics and Technology} 83: 32--44.
\newblock \doi{10.1016/j.infrared.2017.03.009}.

\bibitem[{Zhang et~al.(2015)Zhang, Choi, Daniilidis, Wolf and Kanan}]{vais}
Zhang MM, Choi J, Daniilidis K, Wolf MT and Kanan C (2015) {VAIS: A dataset for
  recognizing maritime imagery in the visible and infrared spectrums}.
\newblock \emph{IEEE Conference on Computer Vision and Pattern Recognition
  Workshops (CVPRW)} : 10--16\doi{10.1109/CVPRW.2015.7301291.}

\bibitem[{Zhao et~al.(2017)Zhao, Shi, Qi, Wang and Jia}]{pspnet}
Zhao H, Shi J, Qi X, Wang X and Jia J (2017) {Pyramid Scene Parsing Network}.
\newblock \emph{IEEE Conference on Computer Vision and Pattern Recognition
  (CVPR)} : 6230--6239\doi{10.1109/CVPR.2017.660}.

\end{thebibliography}

\end{document}